\documentclass[letterpaper]{article} 
\usepackage{aaai25}
\usepackage{times}  
\usepackage{helvet}  
\usepackage{courier}  
\usepackage[hyphens]{url}  
\usepackage{graphicx} 
\usepackage{natbib}  
\usepackage{caption} 
\usepackage{newfloat}
\usepackage{listings}
\DeclareCaptionStyle{ruled}{labelfont=normalfont,labelsep=colon,strut=off} 
\title{Till the Layers Collapse: Compressing a Deep Neural Network through the Lenses of Batch Normalization Layers. }

\author{%
  Zhu Liao\textsuperscript{\rm 1}, Nour Hezbri\textsuperscript{\rm 2}, Victor Qu\'etu\textsuperscript{\rm 1}, Van-Tam Nguyen\textsuperscript{\rm 1}, Enzo Tartaglione\textsuperscript{\rm 1} 
}

\affiliations{
    LTCI, T\'el\'ecom Paris, Institut Polytechnique de Paris, France\\
    \textsuperscript{\rm 1}\texttt{\{name.surname\}@telecom-paris.fr}\\
    \textsuperscript{\rm 2}\texttt{\{name.surname\}@ensae.fr}
}

\usepackage{hyperref}
\usepackage{times}
\usepackage{soul}
\usepackage{url}
\usepackage[utf8]{inputenc}
\usepackage{amsmath}
\usepackage{amsthm}
\usepackage{booktabs}
\usepackage{algorithm, algpseudocode}
\usepackage{enumitem}
\usepackage{colortbl}
\usepackage{tabularx}
\usepackage{rotating}
\usepackage{xcolor}
\usepackage{multirow}
\usepackage{subcaption} 

\usepackage{multirow}
\usepackage{amsfonts}

\usepackage{tikz}

\usepackage{nicematrix}

\usepackage{hhline} 
\usepackage{highlight}
\renewcommand{\opacity}{50}

\newcolumntype{M}{@{}>{\columncolor{white}[0pt][0pt]}c@{}}

\begin{document}

\maketitle

\begin{abstract}

Today, deep neural networks are widely used since they can handle a variety of complex tasks. Their generality makes them very powerful tools in modern technology. However, deep neural networks are often overparameterized. The usage of these large models consumes a lot of computation resources. 
In this paper, we introduce a method called \textbf{T}ill the \textbf{L}ayers \textbf{C}ollapse (TLC), which compresses deep neural networks through the lenses of batch normalization layers. By reducing the depth of these networks, our method decreases deep neural networks' computational requirements and overall latency. We validate our method on popular models such as Swin-T, MobileNet-V2, and RoBERTa, across both image classification and natural language processing (NLP) tasks.

%
\begin{links}
    \link{Code}{https://github.com/ZhuLIAO001/TLC}
\end{links}

\end{abstract}

\section{Introduction}
\label{sec:intro}

Deep neural networks (DNNs) have grown considerably in recent decades, with applications in many different tasks. 
DNNs capture subtle patterns effectively, enabling a wide range of applications. This also allows them to achieve high accuracy.
Their applications cross various domains, including image classification~\cite{barbano2022two}, semantic segmentation~\cite{chaudhry2022lung}, object detection~\cite{carion2020end}, natural language processing~\cite{touvron2023llama}, and the multi-modal tasks~\cite{8292801}. The ability of DNNs to scale with the size of models and datasets has been well-demonstrated~\cite{hestness2017deep}. 

However, while DNNs have shown their scalability, modern DNNs can consist of millions to billions of parameters, which means that the number of floating point operations (FLOPs) required for a single inference is enormous. 
Not only does this require a lot of computing power, but it also creates huge energy consumption and environmental problems. For instance, models like GPT-3~\cite{brown2020language}, which contains 175 billion parameters, have a huge carbon footprint during training, that emphasizes the need for more sustainable AI.

With growing awareness of AI's environmental impact, there are increasing calls for balance. High performance must align with environmental friendliness.
This has led to the rise of model compression techniques. They reduce network size and complexity without impacting performance significantly.
Techniques such as pruning~\cite{lee2018snip,tartaglione2022loss}, which eliminates less critical neurons or weights, and quantization~\cite{han2015deep}, which reduces the precision of weights and activations, have been instrumental in this regard. 
Furthermore, Knowledge distillation~\cite{hinton2015distilling} enables transferring knowledge from large, complex models to smaller, efficient ones.

However, most compression techniques focus on reducing parameters and filters. Few address reducing the model's depth.
Eliminating parameters or filters has relatively little impact on modern computational resources such as GPUs. Indeed, due to the parallel nature of the computation, the size of layers is mainly limited by memory cache and core availability. The main computational bottleneck is the critical path that the forward propagation must pass through~\cite{pmlr-v202-ali-mehmeti-gopel23a}. We would like to specifically minimize this.

This paper addresses this challenge by introducing a novel method, \textbf{T}ill the \textbf{L}ayers \textbf{C}ollapse (TLC), which compresses DNNs looking through the lenses of batch normalization layers. By leveraging batch normalization parameters, TLC identifies and removes less important layers, thereby decreasing computational demands and latency without significantly compromising model performance. Indeed, in rectifier-activated networks, if the standardized signal is mainly positive, we will know that a linear activation would introduce a minimal error during the forward pass. 
Conversely, a mainly negative signal leads to outputs close to zero.
Leveraging this, we can linearize (or remove) layers in the target model that will minimally alter the model's output. 
We empirically validate our approach across image classification and natural language processing (NLP) tasks. It maintains accuracy while improving efficiency.

We summarize, here below, our key messages and contributions.
\begin{itemize}[noitemsep, nolistsep]
    \item We propose a method for evaluating the importance of layers (Sec.~\ref{sec:rank_layer}) based on the value of the batch normalization parameters (Sec.~\ref{sec:remove_layer}).
    \item We propose TLC, a method that identifies and removes redundant layers by leveraging batch normalization parameters (Sec.~\ref{sec:choose_layer}).
    \item 
    TLC is tested across various architectures and datasets. It achieves a balance between reducing layers and maintaining performance (Sec.~\ref{sec:res}).
\end{itemize}

\section{Related Works}
\label{sec:sota}

\begin{figure*}[t]
    \centering
    \includegraphics[width=1.0\textwidth]{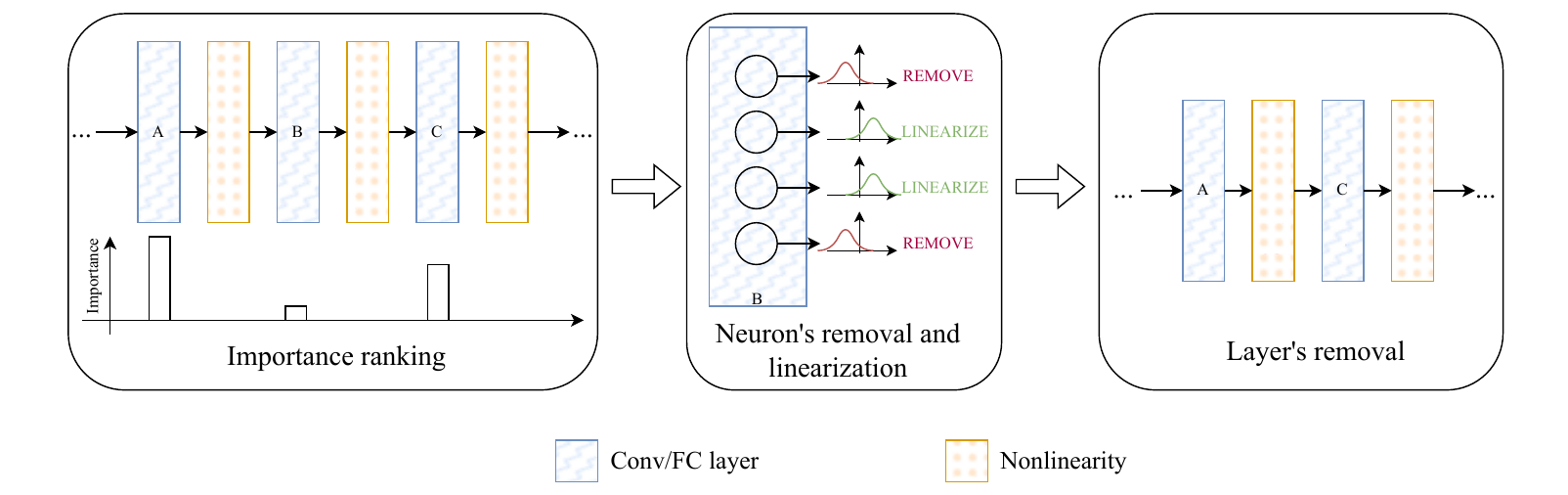}
    \caption{Overview of the key steps for TLC: identification of layer to remove, removal of irrelevant channels, and linearization of the remaining, removal of the layer.
    }
    \label{fig:pipeline}
\end{figure*}

\subsection{Neural Network Depth Reduction}
\label{sec:rela_depth_reduc}

Researchers have been exploring ways to make neural networks shallower without losing their effectiveness.
\cite{8485719} proposed a layer-wise pruning method based on feature representations to shallow deep neural networks, and then retraining the network using knowledge distillation, aiming to reduce network complexity while maintaining performance. This work stated the possibility of designing a layer-based pruning algorithm. 
\cite{pmlr-v202-ali-mehmeti-gopel23a} introduced a channel-wise method to reduce non-linear units while maintaining similar performance.
Moreover, \cite{dror2021layer} proposed a method, Layer Folding (LF), which learns whether non-linear activations can be removed, allowing the folding of consecutive linear layers into one. More specifically, ReLU-activated layers are replaced with PReLU activations and become regularized slopes. 
During post-training, nearly linear PReLUs are removed, and layers are folded. 
Unlike these previous methods, which focus on the activation function level to decide whether it should be linear or non-linear, or analyzing at the feature level to assess the necessity of neurons. Our approach, TLC, directly evaluates the importance of layers and retains only the most essential ones.

\cite{liao2023can} proposed Entropy-Guided Pruning (EGP), which aims to remove entire layers, this method reduces network depth by prioritizing low-entropy layers for pruning.
This method targets layers that are less active and removes them entirely while trying to keep the network's performance stable.
In the same area, \cite{quetu2024simpler} introduced EASIER, a method using entropy-based importance to reduce network depth.
Specifically, EASIER evaluates the importance of different layers within the network and selectively retains the critical layers, thereby simplifying the network structure.
Unlike EGP, which uses unstructured pruning to gradually induce removable layers, often requires multiple training iterations to remove a single layer. And unlike EASIER, which removes one layer after each training. Our approach attempts to remove multiple layers after each training. 
This provides our method with a clear advantage in training efficiency over EGP and EASIER.

\subsection{Layer's Importance Evaluation}
\label{sec:rela_layer_impo}

The evaluation of the layer's importance has become a crucial aspect of model compression, particularly in the last decade. 
\cite{han2015learning} proposed a Weight Magnitude-Based method that assesses neuron importance by analyzing the magnitude of weights. The rationale is that neurons with smaller weights contribute less to the model's output and can be pruned with minimal impact. However, this approach often requires extensive retraining to regain the accuracy lost due to pruning.
\cite{molchanov2016pruning} evaluate the importance of neurons by leveraging gradient information. More specifically, they select neurons to be pruned by using the first-order Taylor expansion to approximate the change in the loss function to estimate layers' importance. 
This method still faces challenges in identifying optimal pruning strategies, especially in very deep networks.

Despite significant progress in layer importance evaluation, balancing complexity reduction and performance remains challenging.
Our TLC method aims to address this challenge by providing an effective layer-evaluating method. 

\subsection{Other BatchNorm-based Pruning Strategies}

Prior studies have used batch normalization statistics to determine filter significance in CNNs for pruning decisions."
For instance, in \cite{NetSlimming}, the scaling parameters of batch normalization layers are used to define a sparsity-inducing penalty during training. After training, these scaling parameters are employed again to identify unimportant channels in the network.
\cite{oh2022batchnormalizationtellsfilter} employs the parameters of batch normalization layers to characterize the pre-activation Gaussian distributions of filters under the assumption of a sufficiently large batch size. 
Filter importance is measured by the expected absolute activation values.
This metric is then used to rank filters, with a specific pruning ratio assigned to each layer based on the degradation in performance caused by pruning the layer in the pre-trained model. 

While our approach shares some similarities with previous works, it differs in two primary aspects. Firstly, our strategy heavily relies on the behavior introduced by the rectifier nonlinearity, and we do not provide a full ranking of filters using batch normalization layers. Instead, we adopt a strict, hard binary ranking (i.e., \emph{ON/OFF} state). Although the pruned neurons within layers may overlap with our \emph{OFF}-state neurons, the \emph{ON}-state neurons in our case are linearized, whereas in \cite{oh2022batchnormalizationtellsfilter} the neurons remain untouched. Secondly, the pruning ratio in \cite{oh2022batchnormalizationtellsfilter} is defined according to the sensitivity of the accuracy to pruning, whereas in our approach, the layer importance metric is defined by the degradation caused by removing the entire layer, which is then leveraged to obtain a ranking of layer removing.

\section{Till the Layers Collapse}
\label{sec:method}
In this section, we introduce our method TLC, which removes complete layers and reduces the depth of the deep neural network with minimal impact on model performance.
We begin by formulating the problem in Sec.~\ref{sec:prb} and conducting an error analysis in Sec.~\ref{sec:error} to explain the motivation behind our approach.
Following, in Sec.~\ref{sec:remove_layer}, we outline our  
surgical layer removal process which leverages the parameters of batch normalization layers. Next, in Sec.~\ref{sec:rank_layer}, we provide an importance ranking of the layers, which will guide the removal of the least significant layers using the aforementioned strategy. Finally, in Sec.~\ref{sec:choose_layer}, we give an overview of the complete pipeline of our method.

\subsection{Problem Formulation}
\label{sec:prb}
Let us consider a DNN consisting of $L$ layers.
Batch norm layers are associated with these neurons such that, in each layer, batch normalization is sandwiched between the affine transformation and the nonlinearity. 
Even in architectures lacking batch norm layers (e.g., transformers), parameters can be estimated via forward propagation and attached.

For the $i$-th neuron of the $l$-th layer, let $x_{l,i}$ denote the output of the affine transformation (e.g., convolution). The batch norm layer applies the following transformation:

\begin{equation}
    \hat{x}_{l,i} = \frac{x_{l,i}- \mu_{l,i}^{B}}{\sqrt{(\sigma_{l,i}^{B})^2 + \epsilon}},~~~~~~~~~z_{l,i} = \gamma_{l,i} \hat{x}_{l,i} + \beta_{l,i} ,
\end{equation}

where $\mu_{l,i}^{B}$ and $\sigma_{l,i}^{B}$ denote the mean and standard deviation of $x_{l,i}$, and $\epsilon$ is an arbitrarily small constant. As we know, $\hat{x}_{l,i}\sim\mathcal{N}(0,1)$. 
$\gamma_{l,i}$ and $\beta_{l,i}$ are learnable parameters that respectively represent the mean and standard deviation of the batch normalization's output: this means that $z_{l,i} \sim \mathcal{N}(\beta_{l,i},\gamma_{l,i}^{2})$.

The bach norm layer is followed by an activation function, where rectifiers are typically used in modern deep neural networks. We denote the rectifier as $\psi_l \text{,}~\forall l \in \left [1,L-1\right ]$.\footnote{Please note that the output layer typically has a different nonlinearity-and besides, it is a layer that can not be removed.} Finally, the output of the neuron is:
\begin{equation}
    \label{eq:activation}
    y_{l,i} = \psi_l(z_{l,i}).
\end{equation}
The distinct feature of rectifiers is that they divide the input space into two regions, with mainly a separate linear function governing each region.
The first linear region is usually where the neuron's output is maintained, or very close as for example in GeLU. If the input of the rectifier is in this region, we regard this neuron as at \emph{ON} state.
The second is where the rectifier's output of the $i$-th neuron is asymptotically zero or negative, but with the output's magnitude being lower for the same input magnitude, as for example in LeakyRelu. This neuron is regarded as at \emph{OFF} state.

To effectively collapse layers, it is important to take into account the different influences of different neurons. Therefore, we propose to dissect the behavior of the layers by analyzing individual neurons, rather than considering the layers as a whole. This refinement approach will be guided by the statistics provided.

\subsection{Effect of the Layer Removal}
\label{sec:error}
\label{sec:remove_layer}
To reduce the depth of DNNs, several works studied the possibility of collapsing the nonlinearity~\cite{dror2021layer}. When collapsing the nonlinearity, a unified behavior across neurons is enforced, forcefully shifting all the neurons to one side of the rectifier.
Herein, we focus on the implications for the neurons when collapsing the nonlinearity, and we aim to determine the error introduced by this process by examining it at the neuron level using the summary statistics provided by the batch normalization layers.

\begin{figure}[t]
    \centering
    \includegraphics[width=1.0\columnwidth]{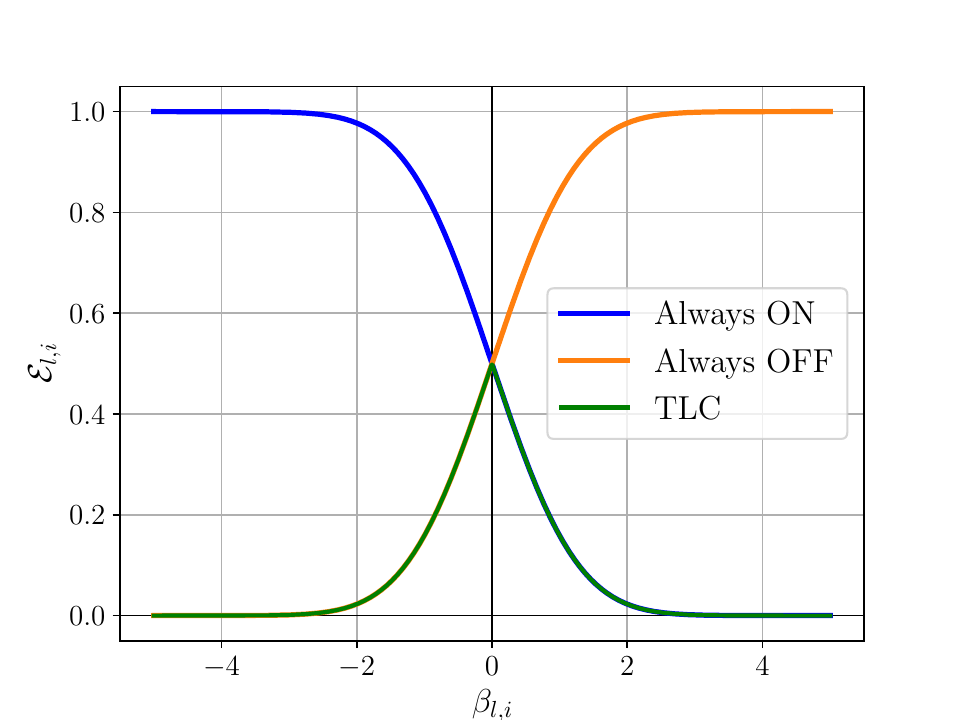}
    \caption{Error plot for the i-th neuron of the l-th layer as a function of the batch norm mean parameter $\beta_{l,i}$ for a standard deviation $\gamma_{l,i}=1$.
    }
    \label{fig:error}
\end{figure}

When we remove a neuron using TLC, we define the likelihood of the error
$\mathcal{E}_{l,i}$ as the probability of a neuron's state being regarded mistakenly. 
When the rectifier $\psi_{l}$ is substituted with the identity function, this is equivalent to shifting all the neurons of the $l$-th layer to the always \emph{ON} state, we call it always ON. 
Conversely, substituting $\psi_{l}$ with the null function pushes all the neurons to the always \emph{OFF} state, we call it always OFF. So, the error likelihood for the $i$-th neuron is:

\begin{equation}
    \label{eq:error_function}
    \mathcal{E}_{l,i}  = \Phi\left(-\frac{|\beta_{l,i}|}{\gamma_{l,i}}\right) ,
\end{equation}

where $\Phi$ denotes the cumulative distribution function of $\mathcal{N}(0,1)$. In Fig.~\ref{fig:error}, the blue curve shows how a neuron's error likelihood varies with $\beta_{l,i}$ when substituting the rectifier with an identity function, and the orange curve shows the error likelihood when substituting the rectifier with a null function. 
Obviously, if all neurons' rectifiers in a layer are substituted uniformly, in either case, there will be unacceptable probability of error.

Based on this analysis,
we devise a layer-removing scheme, in which we selectively linearize the neurons and reconfigure the layers, conditioned on the batch norm layer's parameters. 

Specifically, inside the layer $l$, we discriminate between the \emph{ON} and \emph{OFF} states of the neurons leveraging the computed statistics of the BatchNorm layer $\beta_{l,i}$, $\forall i \in [1, N_l]$.

\begin{itemize}
    \item When $\beta_{l,i}\leq0$, the $i$-th neuron is more likely to be in the \emph{OFF} state, its contribution is therefore marginal and we suppose it can be removed from the network.
    \item When $\beta_{l,i}>0$,  the $i$-th neuron is more likely to be in the \emph{ON} state; it is linearized and merged within the subsequent layer via linear combination 
    similarly to \cite{pilolayerfold}
    to account for its contribution. 
\end{itemize}

\begin{algorithm}[t]
    \caption{Our proposed method TLC.}
    \label{alg:CTBN}
    \begin{algorithmic}[1]
    \Function{{\text{TLC}}($M$, $\mathcal{D}_{\text{train}}$, $\mathcal{D}_{\text{val}}$, $\theta$)}{}
    \State ${M} \leftarrow$ Train$(M, \mathcal{D}_{\text{train}})$ \label{line:train}
    \State $\mathcal{A}^{\text{init}} \leftarrow $Evaluate($M$, $\mathcal{D}_{\text{val}} $) \label{line:evaluate-dense}
    \State $M' \gets M$
    \State $\mathcal{A}_{M'}  \leftarrow \mathcal{A}^{\text{init}}$
    \While{$\mathcal{A}_{M'} \geq \theta \cdot \mathcal{A}^{\text{init}}$}
        \State $M \gets M'$
        \State $L \leftarrow$ list of layers in $M$
        \State $L_{\text{ranked}} \leftarrow \text{Rank}(L)$ \label{line:layer-rank}
        \State $i \leftarrow 1$
        \While{${\mathcal{A}}_{M_{\text{test}}} \geq {\mathcal{A}}_{M'}$} \label{line:remov-ite} \label{line:loss_ex} 
            \State $M' \gets M_{\text{test}}$
            \State $M_{\text{test}} \leftarrow$ $\text{Remove}(M', L_{\text{ranked}}[i])$ \label{line:first_remove} 
            \State $\mathcal{A}_{M_{\text{test}}} \leftarrow \text{Evaluate}(M_{\text{test}}, \mathcal{D}_{\text{val}}) $ \label{line:eva-temp}
            \State $i \leftarrow i + 1$    \label{line:1plus}        
        \EndWhile
        \State ${M'} \leftarrow$ Train$(M', \mathcal{D}_{\text{train}})$ \label{line:retrain}
        \State $\mathcal{A}_{M'} \leftarrow \text{Evaluate}(M', \mathcal{D}_{\text{val}})$  \label{line:loss_retrained}
    \EndWhile
    \State \textbf{return} $M$
    \EndFunction
    \end{algorithmic}
\end{algorithm}

This scheme strikes a balance between two extremes: completely shutting the layer (i.e., always OFF state) and fully linearizing the activation function (i.e., always ON state). 
By adopting this approach, we effectively minimize the error introduced by these two limiting scenarios.
As shown in Fig.~\ref{fig:error}, the overall error likelihood incurred by our scheme (represented by the area under the green curve) is significantly smaller than that resulting from either of the two extreme cases. This demonstrates our approach can remove layers while minimizing the impact on performance.

Please note that in the transformer architectures we adopt in the article, LayerNorm exists before the fully-connected layer, no normalization was implemented between the layer and the activation. 
In this case, LayerNorm parameters were unused. Instead, we calculated the average and standard deviation at the fully connected layer's output.
Based on the computed average, we decide whether to set each neuron to OFF or ON.

\subsection{Layers' Importance Ranking}
\label{sec:rank_layer}

In TLC, we shall remove layers based on the importance ranking we adopt, mainly conditioned on the change in performance the removing of a layer from a complete pre-trained model would engender.

Starting from a pre-trained model $M$, we remove the $l$-th layer with the method we mentioned in Sec.~\ref{sec:remove_layer}. The performance of the pruned model $M_{\text{rem}\{l\}}$ is then evaluated.

We define the layer importance relation $\mathcal{I}$ between the layers $l$ and $l'$ looking at the model's accuracy $\mathcal{A}$ as follows: 
\begin{equation}
    \mathcal{I}(l)<\mathcal{I}(l') \Leftrightarrow  \mathcal{A}(M_{\text{rem}\{l\}}) < \mathcal{A}(M_{\text{rem}\{l'\}}).
\end{equation}

Layers within the model do not affect the model's performance in the same way. Thus, the effects of their removal would vary notably. 
In particular, we expect that removing some layers would result in a drop in performance compared to the original pre-trained model (i.e. $~{\mathcal{A}(M)-\mathcal{A}(M_{\text{rem}\{l\}})>0}$). In other cases though, we might even have that the model's accuracy increases(i.e. $\mathcal{A}(M)-\mathcal{A}(M_{\text{rem}\{l\}})<0$), which can be attributed for example to overfitting of the full model.

\subsection{Overview on TLC}
\label{sec:choose_layer}
In Alg.~\ref{alg:CTBN}, we present our method TLC that strategically leverages batch norm layers to prune layers and reduce the depth of DNNs. 
Given a model $M$, after vanilla training (line~\ref{line:train}), we get model $M'$. We evaluate its initial accuracy $\mathcal{A}^{\text{init}}$ (line~\ref{line:evaluate-dense}) on the validation set $\mathcal{D}_{\text{val}}$, which we use to get an importance ranking of the different layers of the model as in Sec.~\ref{sec:remove_layer} (line~\ref{line:layer-rank}).
Subsequently, we use this ascendent ranking of the layers to guide the layer pruning process, starting with the least important layer, and removing one layer at a time (line~\ref{line:first_remove}). For this step, we first entirely remove the neurons whose average pre-activation is negative ($\beta_{l,i} < 0$) and then the remaining are linearized and fused with the next one, according to Fig.~\ref{fig:pipeline}. The incremental removal of layers redefines each time a new model $M_{\text{test}}$, whose validation accuracy of $M_{\text{test}}$ is iteratively evaluated (line~\ref{line:eva-temp}): if a decrease in the validation accuracy of $M_{\text{test}}$ to $M'$ is detected (line~\ref{line:loss_ex}), the procedure is terminated, and the pruned model $M_{\text{test}}$ substitutes $M'$. Then $M'$ undergoes a retraining process to recover its performance (line~\ref{line:retrain}).
The final model $M$ is the smallest whose accuracy on the validation set does not drop below a relative threshold $\theta$.

\section{Experiments}
\label{sec:result}

\begin{figure}[t]
    \centering
    \includegraphics[width=0.98\columnwidth]{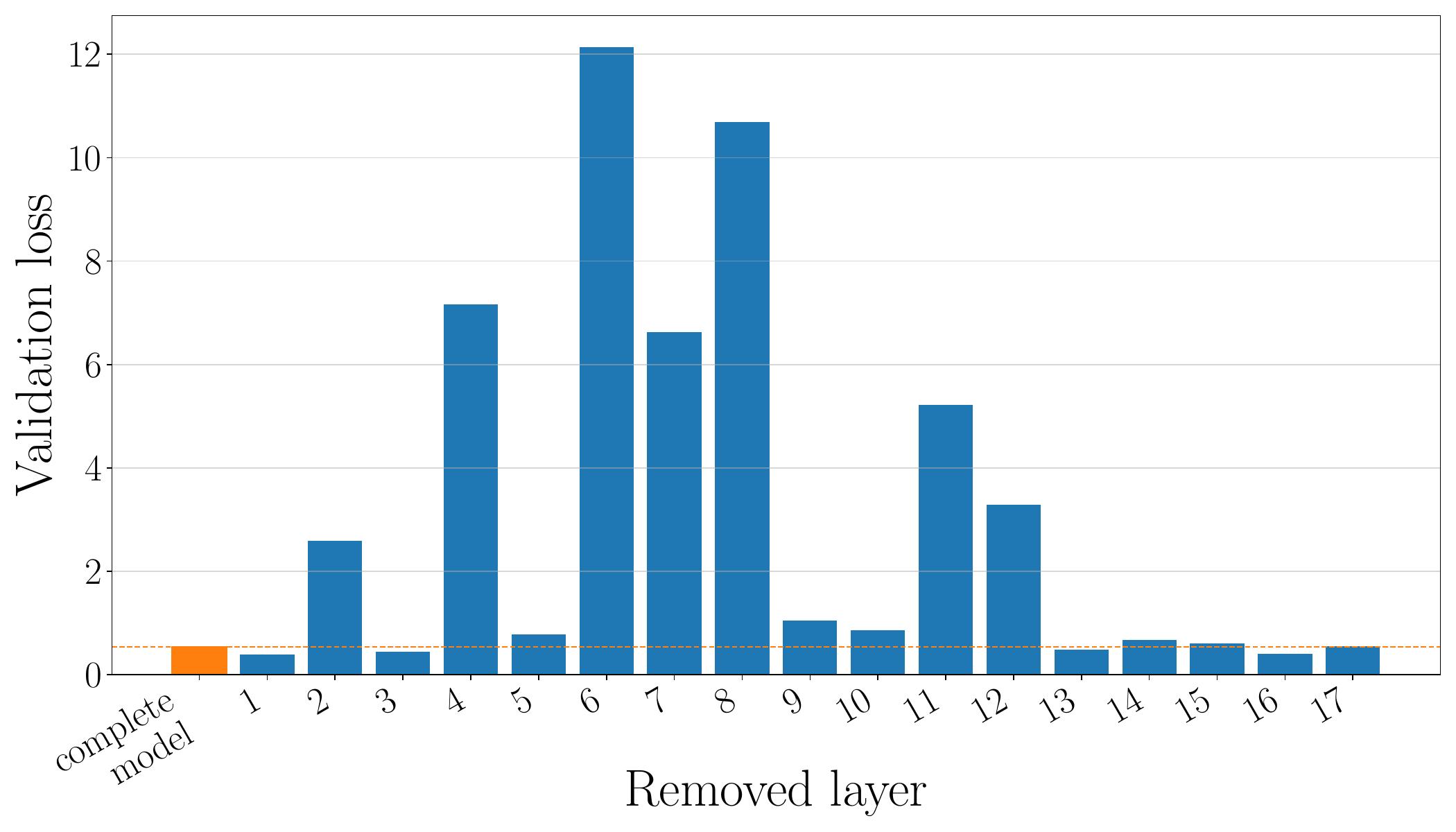}
    \caption{Validation loss for the complete Resnet-18 model pre-trained on Cifar-10 and one layer is removed.
    }
    \label{fig:c10_r18_rank}
\end{figure}

\begin{figure}[t]
    \centering
    \includegraphics[width=0.95\columnwidth]{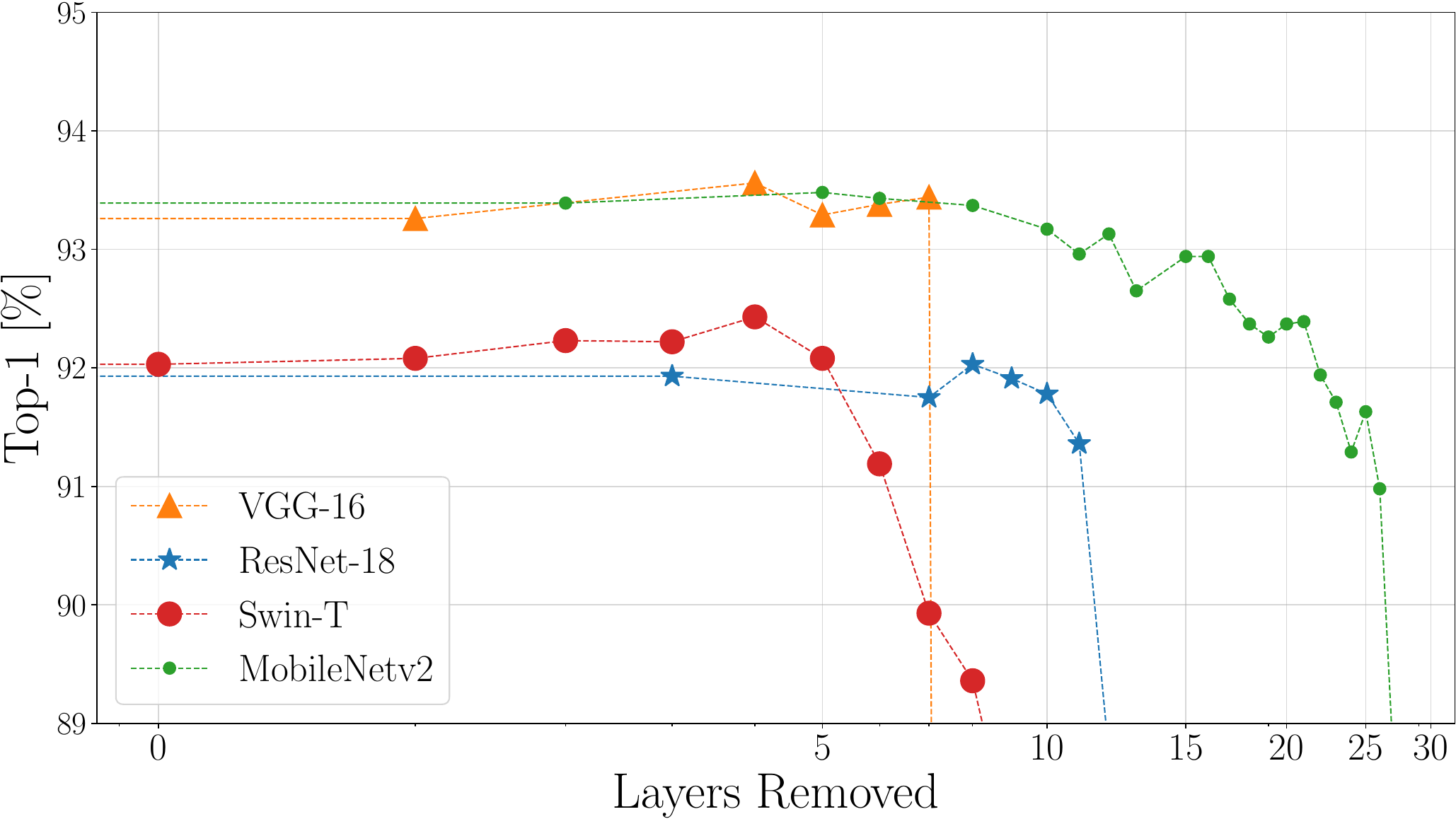}
    \caption{Test performance (top-1) for models trained on CIFAR-10 with different numbers of layers removed by TLC. }
    \label{fig:c10_top1_rem}
\end{figure}

\begin{figure}[t]
    \centering
    \includegraphics[width=1.0\columnwidth]{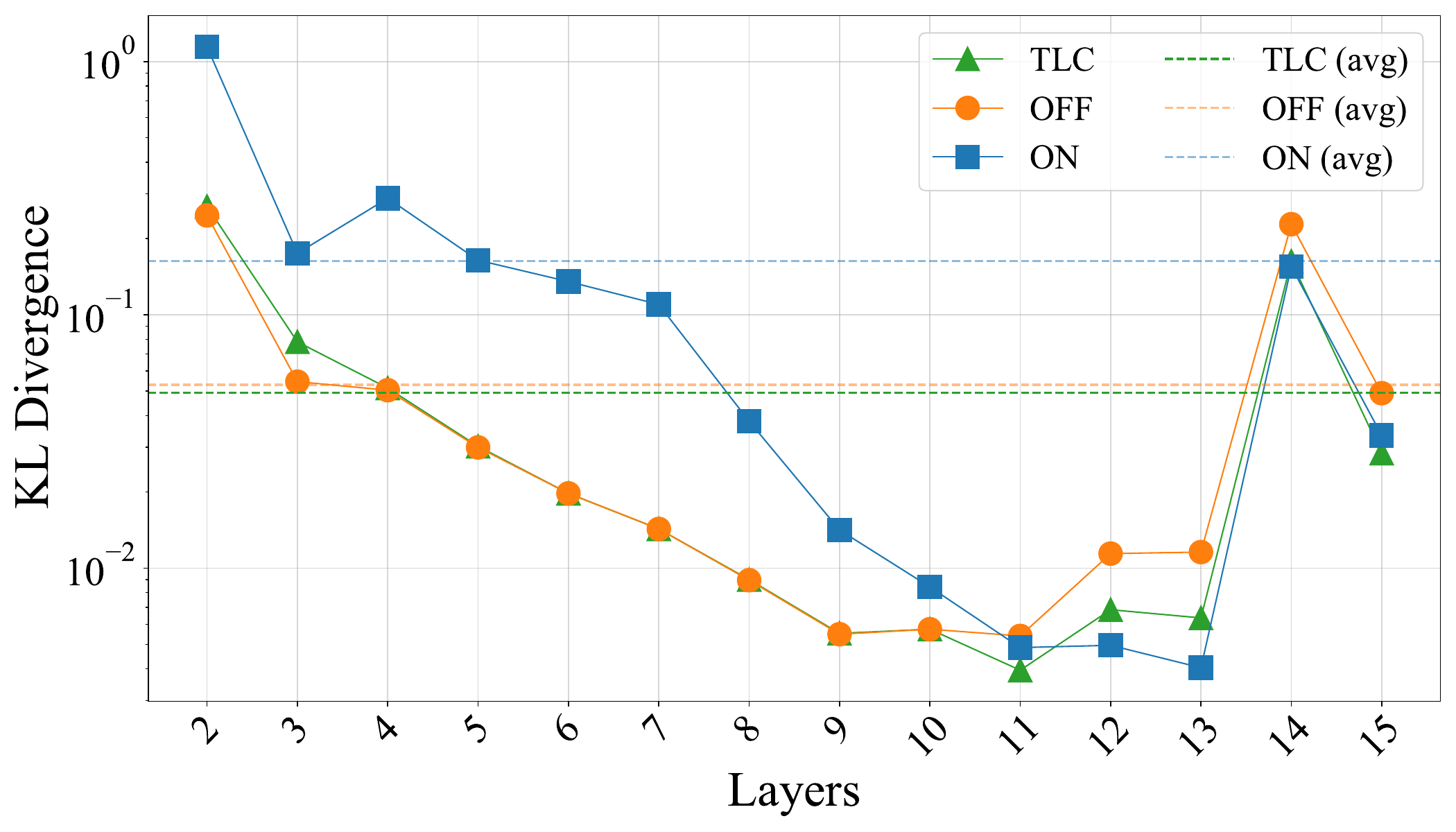}
    \caption{Kullback-Leibler (KL) divergence between the output features of the original VGG-16bn model trained on CIFAR-10 and models removed layers by different methods.
    }
    \label{fig:c10_v16_kl}
\end{figure}

In this section, we evaluate our method on multiple architectures and datasets for image classification and NLP tasks.
All the trainings are performed on an NVIDIA RTX 3090 Ti equipped with 24GB RAM.

\begin{table*}[t]
    \centering
    \resizebox{1.9\columnwidth}{!}{%
    \begin{tabular}{cc|cc|cc|cc|cc}
    \toprule
        \multirow{2}{*}{\textbf{Dataset}} & \multirow{2}{*}{\textbf{Approach}} & \multicolumn{2}{c}{\textbf{ResNet-18}} & \multicolumn{2}{c}{\textbf{~Swin-T~}} & \multicolumn{2}{c}{\textbf{~MobileNet-V2~}} & \multicolumn{2}{c}{\textbf{~VGG-16bn~}} \\
        ~ & ~ & ~~top-1~ & ~Rem.~~ & ~~top-1~ & ~Rem.~~ & ~~top-1~ & ~Rem.~~ & ~~top-1~ & ~Rem.~~ \\
    \toprule
       \multirow{7}{*}{CIFAR-10} & Dense model & 92.00 & 0/17 & 91.63 & 0/12 & 93.64 & 0/35 & 93.09 & 0/15 \\ 
         & Smallest weights   & 88.49 & 11/17 & 86.92 & 3/12 & 10.00 & 1/35 & 90.53 & 7/15 \\
         & Smallest gradients & 88.60 & 11/17 & 86.96 & 3/12 & 10.00 & 1/35 & 90.4 & 7/15 \\
         & EGP & 90.64 & 5/17 & 86.04 & 6/12 & 92.22 & 6/35 & 10.00 & 1/15 \\
         & LF & 90.65 & 1/17 & 85.73 & 2/12 & 89.24 & 9/35 & 86.46 & 1/15 \\ 
         & EASIER & 86.53 & 11/17 & 91.25 & 6/12 & 92.45 & 16/35 & 93.03 & 7/15 \\
         & TLC &  $\bf90.91\pm0.57 $ & \bf 12/17 & $\bf91.98\pm0.07$ & \bf 6/12 & $\bf92.97\pm0.38$ & \bf 17/35 & $\bf93.61\pm0.23$ & \bf 7/15 \\

    \midrule
         \multirow{7}{*}{Tiny-Inet} & Dense model & 41.86 & 0/17 & 75.88 & 0/12 & 45.70 & 0/35 & 58.44 & 0/15 \\
         & Smallest weights   & 37.42 & 8/17 & 72.90 & 1/12 & 0.5 & 1/35 & 56.88 & 1/15 \\
         & Smallest gradients & 37.88 & 8/17 & 72.92 & 1/12 & 0.5 & 1/35 & 57.34 & 1/15 \\   
         & LF & 37.86 & 4/17 & 50.54 & 1/12 & 25.88 & 12/35 & 31.22 &  1/15 \\
         & EGP & 37.44 & 5/17 & 71.48 & 1/12 & 46.88 & 1/35 & --- & --- \\
         & EASIER & 35.84 & 6/17 & 70.94 & 1/12 & 47.58 &   11/35 & 55.16 & 1/15 \\
         & TLC & $\bf38.66\pm0.68$ & \bf 9/17 & \bf $\bf74.07\pm0.02$ & \bf 1/12 &  $\bf47.84\pm0.55$ & \bf 16/35 & \bf $\bf57.63\pm0.65$ &  1/15 \\
    \midrule
       \multirow{7}{*}{PACS} & Dense model & 79.70 & 0/17 & 97.00 & 0/12 & 96.10 & 0/35 & 96.10 & 0/15 \\
         & Smallest weights   & 84.30 & 8/17 & 95.10 & 3/12 & 18.50 & 1/35 & 95.20 & 3/15 \\
         & Smallest gradients & 83.60 & 6/17 & 95.90 & 3/12 & 18.50 & 1/35 & 95.50 & 1/15 \\  
         & LF & 82,90 & 3/17 & 87,70 & 2/12 & 79.70 & 1/35 & 93.60 & 1/15 \\
         & EGP & 81.60 & 3/17 & 93.50 & 4/12 & 17.70 & 3/35 & --- & --- \\
         & EASIER & \bf  88.30 & \bf 9/17 & 93.80 & 3/12 & 94.40 & 7/35 & 95.20 & 3/15 \\
         & TLC &  $84.80\pm0.78$ & 9/17 & \bf  $\bf96.57\pm0.41$ & \bf 4/12 & \bf $\bf94.87\pm0.19$ & \bf 11/35 &  $\bf95.98\pm0.22$ & \bf 4/15 \\
    \midrule
        \multirow{7}{*}{VLCS} & Dense model & 67.85 & 0/17 & 85.83 & 0/12 & 81.83 & 0/35 & 84.62 & 0/15 \\ 
         & Smallest weights   & 65.89 & 16/17 & 69.99 & 5/12 & 6.43 & 1/35 & 80.71 & 7/15 \\
         & Smallest gradients & 66.26 & 11/17 & 70.18 & 5/12 & 6.43 & 1/35 & 80.99 & 7/15 \\  
         & LF & 63.28 & 7/17 & 70.92 & 1/12 & 68.87 & 2/35 &  80.24 & 2/15 \\
         & EGP  & 64.40 & 5/17 & 82.76 & 3/12 & 45.85 & 2/35 & --- & --- \\
         & EASIER & 54.24 & 15/17 & 78.19 & 5/12 & 72.88 & 22/35 & 78.84 & 6/15 \\
         & TLC &  $\bf66.43\pm0.66$ & \bf 16/17 & $\bf82.79\pm0.31$ & \bf 5/12 & $\bf76.11\pm1.18$ & \bf 23/35 & $\bf81.41\pm0.42$ & \bf 7/15 \\
    \midrule
        \multirow{7}{*}{ImageNet} & Dense model & 68.28 & 0/17 & 81.08 & 0/12 & 71.87 & 0/35 & 73.37 & 0/15 \\ 
         & Smallest weights   & 67.80 & 2/17 & 79.74 & 1/12 & 0.1 & 1/35 & 70.67 & 1/15 \\
         & Smallest gradients & 67.56 & 2/17 & 79.71 & 1/12 & 0.1 & 1/35 & 70.12 & 1/15 \\  
         & LF & 67.62 & 1/17 & 73.51 & 1/12 & 7.89 & 1/35 & 72.22 & 2/15 \\
         & EGP  & 61.73 & 2/17 & 78.62 & 1/12 & 0.1 & 1/35 & --- & --- \\
         & EASIER & 67.20 & 2/17 & 78.78 & 1/12 & 41.14 & 2/35 & 1.19 & 1/15 \\
         & TLC & \bf 67.81 & \bf 2/17 & \bf 79.96 & \bf 1/12 & \bf 59.43 & \bf 2/35 & \bf 72.89 & \bf 2/15 \\         
    \bottomrule
    \\
    \end{tabular}
    }
    \caption{Test performance (top-1) and the number of removed layers (Rem.) for all image classification setups considered. 
    The best results between Smallest weights/gradients, LF, EGP, EASIER, and TLC are in \textbf{bold}.
    }
    \label{tab:Image_results}
\end{table*}

\begin{table}[h]
    \centering
    \resizebox{\columnwidth}{!}{%
    \begin{tabular}{cc|cc|cc}
    \toprule
        \multirow{2}{*}{\textbf{Dataset}} & \multirow{2}{*}{\textbf{Approach}} & \multicolumn{2}{c}{\textbf{~BERT~}} & \multicolumn{2}{c}{\textbf{RoBERTa}}  \\
        ~ & ~ & ~~top-1~ & ~Rem.~~ & ~~top-1~ & ~Rem.~~  \\
    \toprule
       \multirow{7}{*}{SST-2} & Dense model & 92.55 & 0/12 & 94.04 & 0/12 \\ 
         & Smallest weights   & 90.14 & 3/12 & 92.20 & 5/12  \\
         & Smallest gradients & 90.25 & 4/12 & 92.43 & 4/12  \\ 
         & LF & 84.52 & 2/12 & 50.92 & 2/12  \\ 
         & EGP & 85.09 & 4/12 & 86.47 & 5/12  \\ 
         & EASIER & 84.63 & 3/12 & 86.81 & 4/12  \\ 
         & TLC & $\bf91.44\pm0.61$ & \bf 4/12 & \bf $\bf93.00\pm0.28$ & \bf 6/12 \\
    \midrule
         \multirow{7}{*}{QNLI} & Dense model & 90.61 & 0/12 & 91.47 & 0/12 \\
         & Smallest weights   & 83.65 & 9/12 & 79.55 & 6/12  \\
         & Smallest gradients & 84.64 & 10/12 &  80.41 & 8/12  \\
         & LF & 49.46 & 1/12 & 50.54 & 2/12  \\ 
         & EGP & 82.85 & 9/12 & 84.66 & 4/12  \\ 
         & EASIER & 50.54 & 3/12 & 50.54 & 3/12  \\          
         & TLC & $\bf84.80\pm0.92$ &  \bf10/12 &  $\bf89.53\pm1.97$ &  \bf 8/12  \\
    \midrule
       \multirow{7}{*}{RTE} & Dense model & 57.04 & 0/12 & 70.40 & 0/12  \\
        & Smallest weights   & 46.93 & 1/12 & \bf 75.81 & \bf 1/12  \\
         & Smallest gradients & 55.23 & 1/12 & 72.20 & 1/12  \\ 
         & LF & 52.71 & 1/12 & 47.29 & 1/12  \\ 
         & EGP & 57.73 & 1/12 & 52.71 &  1/12  \\ 
         & EASIER & 53.07 & 1/12 & 47.29 & 1/12  \\          
         & TLC & $\bf59.08\pm1.68$ & \bf 1/12 & $74.13\pm0.61$  &  1/12  \\
    \bottomrule
    \\
    \end{tabular}
    }
    \caption{Test performance (top-1) and the number of removed layers (Rem.) for all the considered NLP setups. 
    }
    \label{tab:NLP_results}
\end{table}

\subsection{Experimental Setup}
\label{sec:setup}

We assess our method through image classification and NLP tasks. Concerning image classification, our evaluation encompasses four models: ResNet-18~\cite{he2016deep}, MobileNet-V2~\cite{howard2017mobilenets}, VGG-16bn~\cite{simonyan2015deep}, and Swin-T~\cite{liu2021swin}. Models are trained on CIFAR-10~\cite{krizhevsky2009learning}, Tiny-ImageNet~\cite{le2015tiny}, ImageNet dataset~\cite{5206848}, as well as PACS and VLCS from DomainBed~\cite{gulrajani2020search}. 
Training policies follow~\cite{quetu2024dsd2} and~\cite{Xu_2021_CVPR}.

For NLP, our evaluation focuses on two models: BERT~\cite{kenton2019bert} and RoBERTa~\cite{liu2019roberta}. Models are trained on SST-2~\cite{socher2013recursive}, QNLI~\cite{williams2017broad}, and RTE~\cite{bentivogli2009fifth}. We adhere to the training strategies delineated by~\cite{peer2022greedy} for NLP tasks. All the hyperparameters, augmentation strategies, learning policies, and training time are provided in Appendix~\ref{sec:train_detail} and Appendix~\ref{sec:more results}.

We compare our results with the dense model and two additional baselines: removing layers with the lowest weights/gradients. We also compare our method with existing approaches such as EGP~\cite{liao2023can}, Layer folding (LF)~\cite{dror2021layer}, and EASIER~\cite{quetu2024simpler}.

\subsection{Results}
\label{sec:res}

\paragraph{A first overview.} First, we tested our method TLC across different models trained on the CIFAR-10 dataset. 
Fig.~\ref{fig:c10_r18_rank} shows the validation loss of the Resnet-18 complete model trained on CIFAR-10 and the loss after one layer is removed. 
For visualization purposes, we show validation loss here; accuracy plots are in Appendix~\ref{sec:more results}.
The orange bar is the complete model's validation loss, the other bars are the validation losses after the corresponding layers are removed.
The plot shows that removing certain layers can reduce validation loss (equivalently, increases the validation accuracy).

Fig.~\ref{fig:c10_top1_rem} shows top-1 validation trends for four models. Performance remains stable until a critical number of layers are removed, then drops significantly (particularly evident in VGG-16).

To estimate the error at the whole layers' scale, we compare TLC with always OFF and always ON on VGG-16bn trained on CIFAR-10 
using KL divergence between the output features of the original model and each method.
Fig.~\ref{fig:c10_v16_kl} shows TLC yields lower KL divergence (0.049) compared to always ON (0.163) and always OFF (0.053), indicating TLC introduces the least error.

\paragraph{Image classification tasks.}
Table~\ref{tab:Image_results} shows test performance (top-1) and removed layers (Rem.) across all the considered image classification setups.
We discover that removing layers with the lowest sum weights/gradients fails for the MobileNet architecture. Starting with the removal of the first layer, this mechanism tends to focus on removing the last single layer before the classifier head, leading to gradient explosion in subsequent training.
Moreover, EGP results on the VGG-16bn architecture are reported only for CIFAR-10 due to the layer collapse phenomenon: when forcing a layer to have zero entropy, it remains in the \emph{OFF} state; this prevents signal transmission. 
Accordingly, to save computational resources, we did not train VGG-16bn on other datasets with EGP. 
However, architectures such as ResNet-18, Swin-T, and MobileNet-V2 do not exhibit this problem due to the presence of skip connections, which provide alternate paths for signal flow even if an entire layer is pruned.
Meanwhile, 
TLC retains enough \emph{ON} state neurons to ensure proper signal transmission.
Moreover, 
TLC avoids removing performance-critical layers.
As a result, TLC does not exhibit the problems that removing layers with the lowest sum weights/gradients and EGP encountered, it works well with all considered architectures.

Compared to LF, TLC removes more layers while maintaining or improving top-1 accuracy.
Compared to EASIER, TLC achieves comparable (in most cases better) results for models trained on CIFAR-10, Tiny-ImageNet, and PACS, as well as for ResNet-18 and Swin-T models trained on ImageNet. However, for models trained on VLCS, and for MobileNet-V2 and VGG-16bn models trained on ImageNet, TLC consistently yields significantly better top-1 accuracy at the same level of layer removal. Moreover, EASIER is an iterative method, it removes only one layer at a time. Since TLC tries to remove multiple layers together, our method has a significant advantage in training efficiency.

\paragraph{NLP tasks.}
Table~\ref{tab:NLP_results} shows the results for all the NLP setups. 
Similarly to what observed for image classification tasks, TLC can obtain models with layer removal and maintain good performance. 
The results show that removing the layer with the lowest sum of weights/gradients results
performs close to 
TLC in most setups. Both methods can remove layers from models with minimal or no performance degradation. This reveals the presence of redundancy in these models.
It also appears that On NLP tasks, TLC outperforms LF, EGP, and EASIER by achieving higher accuracy, more removable layers, or both.
The exception rises for RTE, where in general the number of removable layers is low. We hypothesize that the pre-trained models are not a good fit for this specific downstream task, also looking at a lower performance of the Dense model compared to SST-2 or QNLI: for this, removing layers might not be a viable strategy. This raises a warning when employing approaches that reduce the model's depth.

\subsection{Ablation Study}
\label{sec:ablation_study}

Table~\ref{tab:resnet18_acti} shows the test performance of ResNet-18 on CIFAR-10, for different rectifiers versus the number of linearized layers. As expected, TLC is compatible with the most common rectifiers, removing a similar amount of layers. The performance gap recorded is due to the different nonlinearities employed. It appears that TLC is not bound to a specific one and is effective with all the most popular choices.

Table~\ref{tab:flops_time} presents a measure of FLOPs and real memory occupation, on Swin-T trained on CIFAR-10 with layers collapsed through TLC. 
Generally, the fewer layers the network has, the smaller the number of FLOPs, and the smaller the memory usage.

\begin{table}[t]
    \centering
    \resizebox{0.7\columnwidth}{!}
    {
    \begin{tabular}{cccc}
    \toprule
    \bf Activation              &\bf Approach    & \bf top-1 & \bf Rem. \\ \midrule
                                & Dense model        & 92.00     & 0/17  \\
    \multirow{-2}{*}{ReLU}      & TLC     & 91.36     & 12/17  \\ \midrule
                                & Dense model        & 92.22     & 0/17  \\
    \multirow{-2}{*}{SiLU}      & TLC     & 91.72     & 12/17  \\ \midrule
                                & Dense model        & 91.55     & 0/17  \\
    \multirow{-2}{*}{PReLU}     & TLC     & 90.57     & 12/17  \\ \midrule
                                & Dense model        & 91.79     & 0/17  \\
    \multirow{-2}{*}{LeakyReLU} & TLC     & 92.00     & 11/17  \\  \midrule
                                & Dense model        & 91.83     & 0/17  \\
    \multirow{-2}{*}{GELU}      & TLC     & 91.84     & 12/17   \\ 
    \bottomrule
    \end{tabular}
    }
    \caption{Analysis with different activation functions on ResNet-18 trained on CIFAR-10.}
    \label{tab:resnet18_acti}
\end{table}

\begin{table}[h]
    \centering
    \resizebox{0.7\columnwidth}{!}
    {%
    \begin{tabular}{cccc}
    \toprule
        \multirow{2}{*}{\bf Rem.} & \multirow{2}{*}{\bf MFLOPs} & {\bf Mem.usage}  &\multirow{2}{*}{\bf top-1}  \\ 
        \bf  & \bf  & {\bf [MBs]} &  \\
    \midrule
        0/12 & 8987.13  & 115.80  & 91.63\\ 
        1/12 & 8582.51  & 102.54  & 92.03\\ 
        2/12 & 8177.89  & 100.35  & 92.08\\ 
        3/12 & 7773.27  & 83.40  & 92.23\\ 
        4/12 & 7368.65  & 67.33  & 92.22\\ 
        5/12 & 6964.03  & 63.95  & 92.43\\ 
        6/12 & 6559.41  & 59.44  & 92.08\\ 
        7/12 & 6154.79  & 58.45  & 91.19\\ 
        8/12 & 5750.18  & 57.47   & 89.93\\ 

    \bottomrule
    \\
    \end{tabular}
    }
    \caption{MFLOPs and Memory usage [MBs] of Swin-T on CIFAR-10 on NVIDIA RTX 4500.}
    \label{tab:flops_time}
\end{table}

\begin{table}[t]
    \centering
    \resizebox{0.85\columnwidth}{!}
    {
    \begin{tabular}{cccc}
    \toprule
    \bf Model              &\bf Approach    & \bf top-1 & \bf Rem. \\ \midrule
                                & Dense model        & 92.00     & 0/17  \\
    \multirow{-2}{*}{ResNet-18}      & TLC-finetuning     & 91.12     & 7/17  \\ \midrule
                                & Dense model        & 91.63     & 0/12  \\
    \multirow{-2}{*}{Swin-T}      & TLC-finetuning      & 89.49     & 2/12  \\ \midrule
                                & Dense model        & 93.64     & 0/35  \\
    \multirow{-2}{*}{MobileNet-V2}     & TLC-finetuning      & 92.52    & 15/35  \\ \midrule
                                & Dense model        & 93.09     & 0/15  \\
    \multirow{-2}{*}{VGG-16bn}      & TLC-finetuning      & 92.08     & 8/15   \\ 
    \bottomrule
    \end{tabular}
    }
    \caption{
    Analysis for models trained on CIFAR-10 dataset and pruned by the TLC-finetuning method.
    }
    \label{tab:finetune_TLC}
\end{table}

\subsection{Limitations}
\label{sec:limitations}
TLC is a successful approach to alleviate deep neural networks' computational burden by decreasing their depth. Meanwhile, we also notice that TLC leads to long training times and high computational requirements.

To reduce training costs, we propose TLC-finetuning, an approach with a shorter fine-tuning process that focuses on the final training stage.
We tested the TLC-finetuning method on different models with the CIFAR-10 dataset. As shown in Table~\ref{tab:finetune_TLC}, although the ability to remove layers is not as significant as the method that involves full retraining in each iteration, TLC-finetuning still produces models that retain good top-1 performance while allowing for layer removing. It appears that our method can be scaled to larger language models and remains effective in more complex scenarios. We leave further exploration and refinement of this approach for future research.

\section{Conclusion}
\label{sec:conclusion}

In this work, we have presented TLC, a method designed to reduce the depth of DNNs efficiently. By utilizing the parameters from batch normalization layers, TLC can identify and remove less critical layers while maintaining a good model performance. Our experiments across multiple image classification and NLP tasks demonstrate the robustness and effectiveness of TLC compared to existing methods.

TLC is a step forward in the quest for more sustainable and efficient neural networks, and we hope that in the future we will find even more efficient and environmentally friendly AI.

\newpage
\clearpage 

\section*{Acknowledgments}
This work was supported by several funding bodies. Part of the work was funded by the Hi!PARIS Center on Data Analytics and Artificial Intelligence. It also received support from the European Union’s HORIZON Research and Innovation Programme under grant agreement No. 101120657, as part of the ENFIELD project (European Lighthouse to Manifest Trustworthy and Green AI). Additionally, funding was provided by the French National Research Agency (ANR) under grant agreements ANR-22-PEFT-0003 and ANR-22-PEFT-0007, as part of the France 2030 initiative, specifically the NF-NAI and NF-FITNESS projects. The project also received funding from the European Union’s Horizon Europe Research and Innovation Programme under grant agreement No. 101120237 (ELIAS). Zhu Liao acknowledges financial support from the China Scholarship Council (CSC).

\bibliography{main}

\appendix
\newpage
\clearpage  
\newpage
\section{Details on the Learning Strategies Employed}
\label{sec:train_detail}

The implementation details used in this paper are presented here.

CIFAR-10 is augmented with per-channel normalization, random horizontal flipping, and random shifting by up to four pixels in any direction.
For the datasets of DomainBed, the images are augmented with per-channel normalization, random horizontal flipping, random cropping, and resizing to 224. The brightness, contrast, saturation, and hue are also randomly affected with a factor fixed to 0.4.
Tiny ImageNet is augmented with per-channel normalization and random horizontal flipping.
ImageNet is augmented with per-channel normalization, random horizontal flipping, random cropping, and resizing to 224. 
The sequence length of SST-2, QNLI, and RTE is set to 128.

All weights from ReLU-activated layers are set as prunable for ResNet-18 and VGG-16bn. For Swin-T, BERT, and RoBERTa, all weights from GELU-activated layers are prunable. while for MobileNetv2 all weights from ReLU6-activated layers are considered in the pruning.
Neither biases nor batch normalization parameters are pruned. 

The training hyperparameters used in the Image classification experiments are presented in Table~\ref{tab:learning_strategies}, hyperparameters used in the NLP experiments are presented in Table~\ref{tab:NLP_learning_strategies}. 
Our code is attached to this supplementary material and will be publicly available upon acceptance of the article.

\begin{table*}[h!]
    \centering
    \resizebox{2.0\columnwidth}{!}{
    \begin{tabular}{c c c c c c c c c c}
        \toprule
        \bf Model & \bf Dataset & \bf Epochs & \bf Batch & \bf Opt. & \bf Mom. & \bf LR & \bf Milestones & \bf Drop Factor & \bf Weight Decay \\
        \midrule
         ResNet-18 & CIFAR-10 & 160 & 128 & SGD & 0.9 & 0.1 & [80, 120] & 0.1 & 1e-4 \\
         Swin-T & CIFAR-10 & 160 & 128 & SGD & 0.9 & 0.001 & [80, 120] & 0.1 & 1e-4\\
         MobileNetv2 & CIFAR-10 & 160 & 128 & SGD & 0.9 & 0.1 & [80, 120] & 0.1 & 1e-4\\
         VGG-16bn & CIFAR-10 & 160 & 128 & SGD & 0.9 & 0.1 & [80, 120] & 0.1 & 1e-4\\      
         ResNet-18 & PACS & 30 & 16 & SGD & 0.9 & 0.001 & [24] & 0.1 & 5e-4 \\
         Swin-T & PACS & 30 & 16 & SGD & 0.9 & 0.001 & [24] & 0.1 & 5e-4 \\
         MobileNetv2 & PACS & 30 & 16 & SGD & 0.9 & 0.001 & [24] & 0.1 & 5e-4 \\
         VGG-16bn & PACS & 30 & 16 & SGD & 0.9 & 0.001 & [24] & 0.1 & 5e-4 \\
         ResNet-18 & VLCS & 30 & 16 & SGD & 0.9 & 0.001 & [24] & 0.1 & 5e-4 \\
         Swin-T & VLCS & 30 & 16 & SGD & 0.9 & 0.001 & [24] & 0.1 & 5e-4 \\
         MobileNetv2 & VLCS & 30 & 16 & SGD & 0.9 & 0.001 & [24] & 0.1 & 5e-4 \\
         VGG-16bn & VLCS & 30 & 16 & SGD & 0.9 & 0.001 & [24] & 0.1 & 5e-4 \\
         ResNet-18 & Tiny ImageNet & 160 & 128 & SGD & 0.9 & 0.1 & [80, 120] & 0.1 & 1e-4 \\
         Swin-T & Tiny ImageNet & 160 & 128 & SGD & 0.9 & 0.001 & [80, 120] & 0.1 & 1e-4 \\
         MobileNetv2 & Tiny ImageNet & 160 & 128 & SGD & 0.9 & 0.1 & [80, 120] & 0.1 & 1e-4 \\
         VGG-16bn & Tiny ImageNet & 160 & 128 & SGD & 0.9 & 0.1 & [80, 120] & 0.1 & 1e-4 \\
         ResNet-18 & ImageNet & 90 & 128 & SGD & 0.9 & 0.1 & [30, 60] & 0.1 & 1e-4 \\
         Swin-T & ImageNet & 90 & 128 & SGD & 0.9 & 0.001 & [30, 60] & 0.1 & 1e-4 \\
         MobileNetv2 & ImageNet & 90 & 128 & SGD & 0.9 & 0.1 & [30, 60] & 0.1 & 1e-4 \\
         VGG-16bn & ImageNet & 90 & 128 & SGD & 0.9 & 0.1 & [30, 60] & 0.1 & 1e-4 \\         
          \bottomrule
    \end{tabular}}
    \caption{Table of the different employed learning strategies for image classification tasks.}
    \label{tab:learning_strategies}
\end{table*}

\begin{table*}[h!]
    \centering
    \begin{tabular}{c c c c c c c c c}
        \toprule
        \bf Model & \bf Dataset & \bf Epochs & \bf Batch & \bf Opt.  & \bf LR & $\bf\beta_{1}$ & \bf $\bf\beta_{2}$  & $\bf\epsilon$\\
        \midrule
         BERT & QNLI & 3 & 32 & AdamW & 2e-5 & 0.9 & 0.999 & 1e-8 \\
         RoBERTa & QNLI & 3 & 32 & AdamW & 2e-5 & 0.9 & 0.999 & 1e-8\\
         BERT & RTE & 3 & 32 & AdamW & 2e-5 & 0.9 & 0.999 & 1e-8\\
         RoBERTa & RTE & 3 & 32 & AdamW & 2e-5 & 0.9 & 0.999 & 1e-8\\
         BERT & SST-2 & 3 & 32 & AdamW & 2e-5 & 0.9 & 0.999 & 1e-8\\
         RoBERTa & SST-2 & 3 & 32 & AdamW & 2e-5 & 0.9 & 0.999 & 1e-8\\
          \bottomrule
    \end{tabular}
    \caption{Table of the different employed learning strategies for NLP tasks.}
    \label{tab:NLP_learning_strategies}
\end{table*}

\section{Detailed Results}
\label{sec:more results}
\begin{figure}[t]
    \centering
    \includegraphics[width=1.0\columnwidth]{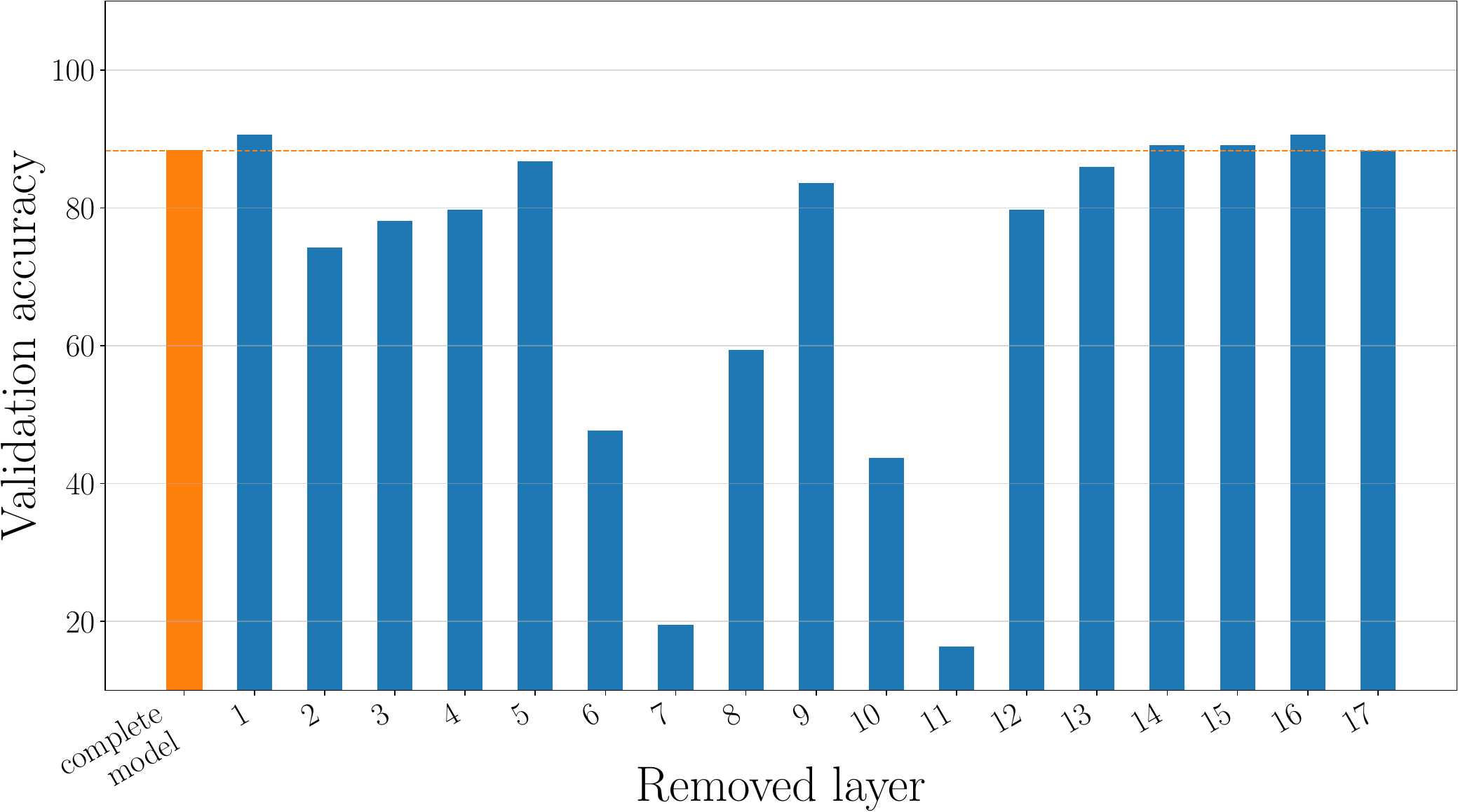}
    \caption{Validation accuracy for the complete ResNet-18 model pretrained on CIFAR-10 and one layer is removed.}
    \label{fig:c10_r18_acc_rank}
\end{figure}

Figure~\ref{fig:c10_r18_acc_rank} shows the validation accuracy of the ResNet-18 complete model trained on CIFAR-10 and the accuracy after one layer is removed. The orange bar is the complete model's validation accuracy, the other bars are the validation accuracies after the corresponding layers are removed from the complete model.
It is surprising that after some layers are removed, the model even yields better performances.

The training time of all the TLC results are presented in Table~\ref{tab:training_time} and Table~\ref{tab:NLP_training_time}. 
All the trainings are performed on an NVIDIA RTX 3090 Ti equipped with 24GB RAM.
It appears that in most cases, TLC demonstrates a significant efficiency advantage over removing layers with the lowest weights/gradients, EGP, and EASIER in generating removable layers.
For example, when training ResNet18 on CIFAR-10, TLC produces one more removable layer than EASIER and smallest wights/gradients while reducing the training time by more than 4 hours. Although the training time of EGP is comparable to that of TLC, TLC yields 7 more removable layers and a better performance.

\begin{table*}[!ht]
    \centering
    \resizebox{2\columnwidth}{!}{%
    \begin{tabular}{cc|ccc|ccc|ccc|ccc}
    \toprule
        \multirow{2}{*}{\textbf{Dataset}} & \multirow{2}{*}{\textbf{Approach}} & \multicolumn{3}{c}{\textbf{ResNet-18}} & \multicolumn{3}{c}{\textbf{~Swin-T~}} & \multicolumn{3}{c}{\textbf{~MobileNet-V2~}} & \multicolumn{3}{c}{\textbf{~VGG-16bn~}} \\
        ~ & ~ & ~~top-1~ & ~Rem.~~ &Time & ~~top-1~ & ~Rem.~~ &Time & ~~top-1~ & ~Rem.~~ &Time & ~~top-1~ & ~Rem.~~ &Time \\
    \toprule
       \multirow{7}{*}{CIFAR-10} & Dense model & 92.00 & 0/17 & 0h48 & 91.63 & 0/12 & 1h53 & 93.64 & 0/35 & 0h43 & 93.09 & 0/15 & 1h01 \\ 
         & Smallest weights   & 88.49 & 11/17 & 9h37 & 86.92 & 3/12 & 7h32 & 10.00 & 1/35 & 1h27 & 90.53 & 7/15 & 8h09 \\
         & Smallest gradients & 88.60 & 11/17 & 9h36 & 86.96 & 3/12 & 7h33 & 10.00 & 1/35 & 1h27 & 90.4 & 7/15 & 8h09 \\
         & LF & 90.65 & 1/17 & 2h31 & 85.73 & 2/12 & 6h11 & 89.24 & 9/35 & 8h49 & 86.46 & 1/15 & 1h59 \\ 
          & EGP & 90.64 & 5/17 & 5h41 & 86.04 & 6/12 & 7h34 & 92.22 & 6/35 & 11h12 & 10.00 & 1/15 & 2h03 \\
         & EASIER & 86.53 & 11/17 & 9h47 & 91.25 & 6/12 & 13h11 & 92.45 & 16/35 & 12h28 & 93.03 & 7/15 & 8h12 \\
         & TLC &  91.36 & 12/17 & 5h36 & 92.08 & 6/12 & 13h10 & 92.94 &  17/35 & 7h54 & 93.44 &  7/15 & 5h07 \\

    \midrule
         \multirow{7}{*}{Tiny-Inet} & Dense model & 41.86 & 0/17 & 2h15 & 75.88 & 0/12 & 5h41 & 45.70 & 0/35 & 1h47 & 58.44 & 0/15 & 3h14 \\
         & Smallest weights   & 37.42 & 8/17 & 20h15 & 72.90 & 1/12 & 11h23 & 0.5 & 1/35 & 3h34 & 56.88 & 1/15 & 6h27 \\
         & Smallest gradients & 37.88 & 8/17 & 20h13 & 72.92 & 1/12 & 11h23 & 0.5 & 1/35 & 3h34 & 57.34 & 1/15 & 6h27 \\   
         & LF & 37.86 & 4/17 & 3h40 & 50.54 & 1/12 & 20h42 & 25.88 & 12/35 & 1d14h12 & 31.22 &  1/15 & 2d22h18 \\
         & EGP & 37.44 & 5/17 & 13h37 & 71.48 & 1/12 & 11h24 & 46.88 & 1/35 & 5h23 & --- & --- & --- \\
         & EASIER & 35.84 & 6/17 & 15h45 & 70.94 & 1/12 & 11h24 & 47.58 &   11/35 & 21h36 & 55.16 & 1/15 & 6h28 \\
         & TLC & 39.6 & 9/17 & 18h03 & 74.10 & 1/12 & 11h22 &  48.24 & 16/35 & 16h12 & 58.20 &  1/15 & 6h28 \\
    \midrule
       \multirow{7}{*}{PACS} & Dense model & 79.70 & 0/17 & 0h46 & 97.00 & 0/12 & 0h57 & 96.10 & 0/35 & 0h34 & 96.10 & 0/15 & 1h20 \\
         & Smallest weights   & 84.30 & 8/17 & 6h54 & 95.10 & 3/12 & 3h48 & 18.50 & 1/35 & 1h09 & 95.20 & 3/15 & 5h21 \\
         & Smallest gradients & 83.60 & 6/17 & 5h22 & 95.90 & 3/12 & 3h49 & 18.50 & 1/35 & 1h09 & 95.50 & 1/15 & 2h40 \\  
         & LF & 82,90 & 3/17 & 1h49 & 87,70 & 2/12 & 4h41 & 79.70 & 1/35 & 4h15 & 93.60 & 1/15 & 2h16 \\
         & EGP & 81.60 & 3/17 & 5h26 & 93.50 & 4/12 & 2h54 & 17.70 & 3/35 & 1h11 & ---  & --- & ---\\
         & EASIER &  88.30 & 9/17 & 7h51 & 93.80 & 3/12 & 3h51 & 94.40 & 7/35 & 4h41 & 95.20 & 3/15 & 5h22 \\
         & TLC &  85.90 & 9/17 & 6h59 & 97.10 & 4/12 & 3h50 & 95.00 &  11/35 & 2h54 & 95.90 & 4/15 & 6h42 \\
    \midrule
        \multirow{7}{*}{VLCS} & Dense model & 67.85 & 0/17 & 3h49 & 85.83 & 0/12 & 2h22 & 81.83 & 0/35 & 3h00 & 84.62 & 0/15 & 2h31 \\ 
         & Smallest weights   & 65.89 & 16/17 & 64h53 & 69.99 & 5/12 & 14h13 & 6.43 & 1/35 & 6h01 & 80.71 & 7/15 & 20h08 \\
         & Smallest gradients & 66.26 & 11/17 & 45h48 & 70.18 & 5/12 & 14h12 & 6.43 & 1/35 & 6h01 & 80.99 & 7/15 & 20h08 \\  
         & LF & 63.28 & 7/17 & 4h28 & 70.92 & 1/12 & 4h44 & 68.87 & 2/35 & 7h12 &  80.24 & 2/15 & 4h37 \\
         & EGP  & 64.40 & 5/17 & 53h43 & 82.76 & 3/12 & 7h11 & 45.85 & 2/35 & 6h02 & --- & ---  & ---\\
         & EASIER & 54.24 & 15/17 & 61h19 & 78.19 & 5/12 & 14h18 & 72.88 & 22/35 & 69h37 & 78.84 & 6/15 & 17h52 \\
         & TLC &  65.98 &  16/17 & 34h21 & 82.48 & 5/12 & 9h28 & 77.63 & 23/35 & 33h09 & 81.83 & 7/15 & 12h41 \\
    \midrule
        \multirow{7}{*}{ImageNet} & Dense model & 68.28 & 0/17 & 3d02h47 & 81.08 & 0/12 & 9d02h15 & 71.87 & 0/35 & 3d08h21 & 73.37 & 0/15 & 8d12h19 \\ 
         & Smallest weights   & 67.80 & 2/17 & 9d08h23 & 79.74 & 1/12 & 18d04h29 & 0.1 & 1/35 & 6d16h41 & 70.67 & 1/15 & 17d00h38 \\
         & Smallest gradients & 67.56 & 2/17 & 9d08h22 & 79.71 & 1/12 & 18d04h30 & 0.1 & 1/35 & 6d16h42 & 70.12 & 1/15 & 17d00h42 \\  
         & LF & 67.62 & 1/17 & 7d02h37 & 73.51 & 1/12 & 30d09h07 & 7.89 & 1/35 & 12d22h18 & 72.22 & 2/15 & 16d10h51 \\
         & EGP  & 61.73 & 2/17 & 9d08h31 & 78.62 & 1/12 & 18d04h35 & 0.1 & 1/35 & 10d01h22 & --- & --- & --- \\
         & EASIER & 67.20 & 2/17 & 9d08h29 & 78.78 & 1/12 & 18d04h35 & 41.14 & 2/35 & 10d01h12 & 1.19 & 1/15 & 17d01h13 \\
         & TLC & 67.81 & 2/17 & 9d08h24 & 79.96 &  1/12 & 18d04h34 &  59.43 &  2/35 & 10d01h08 &  72.89 &  2/15 & 25d12h57 \\         
    \bottomrule
    \\
    \end{tabular}
    }
    \caption{Test performance (top-1), the number of removed layers (Rem.) and training time for all the considered image classification setups.}
    \label{tab:training_time}
\end{table*}

\begin{table*}[!ht]
    \centering
    \resizebox{1.4\columnwidth}{!}{%
    \begin{tabular}{cc|ccc|ccc}
    \toprule
        \multirow{2}{*}{\textbf{Dataset}} & \multirow{2}{*}{\textbf{Approach}} & \multicolumn{3}{c}{\textbf{~BERT~}} & \multicolumn{3}{c}{\textbf{RoBERTa}}  \\
        ~ & ~ & ~~top-1~ & ~Rem.~~ &Time & ~~top-1~ & ~Rem.~~ &Time  \\
    \toprule
       \multirow{7}{*}{SST-2} & Dense model & 92.55 & 0/12 & 0h21 & 94.04 & 0/12 & 0h21 \\ 
         & Smallest weights   & 90.14 & 3/12 & 1h24 & 92.20 & 5/12 & 2h06  \\
         & Smallest gradients & 90.25 & 4/12 & 1h45 & 92.43 & 4/12 & 1h45  \\ 
         & LF & 84.52 & 2/12 & 3h49 & 50.92 & 2/12 & 2h25  \\ 
         & EGP & 85.09 & 4/12 & 0h42 & 86.47 & 5/12 & 0h42  \\ 
         & EASIER & 84.63 & 3/12 & 1h24 & 86.81 & 4/12 & 1h45  \\ 
         & TLC & 91.40 & 4/12 & 0h42 & 93.00 & 6/12 & 0h42 \\
    \midrule
         \multirow{7}{*}{QNLI} & Dense model & 90.61 & 0/12 & 0h34 & 91.47 & 0/12 & 0h35 \\
         & Smallest weights   & 83.65 & 9/12 & 5h40 & 79.55 & 6/12 & 4h05  \\
         & Smallest gradients & 84.64 & 10/12 & 6h14 &  80.41 & 8/12 & 5h15  \\
         & LF & 49.46 & 1/12 & 2h39 & 50.54 & 2/12 & 2h13  \\ 
         & EGP & 82.85 & 9/12 &5h40 & 84.66 & 4/12 & 1h10  \\ 
         & EASIER & 50.54 & 3/12 & 2h16 & 50.54 & 3/12  & 2h20 \\          
         & TLC & 84.80 &  10/12  &2h50 &  90.50 &  8/12 & 3h30  \\
    \midrule
       \multirow{7}{*}{RTE} & Dense model & 57.04 & 0/12 & 0h02 & 70.40 & 0/12 & 0h03  \\
        & Smallest weights   & 46.93 & 1/12 &0h05 & 75.81 & 1/12 & 0h06  \\
         & Smallest gradients & 55.23 & 1/12  &0h05 & 72.20 & 1/12 & 0h06  \\ 
         & LF & 52.71 & 1/12 &0h05 & 47.29 & 1/12 & 0h09  \\ 
         & EGP & 57.73 & 1/12 &0h05 & 52.71 &  1/12  & 0h06 \\ 
         & EASIER & 53.07 & 1/12 &0h05 & 47.29 & 1/12 & 0h06  \\          
         & TLC & 61.37 & 1/12 &0h05 & 74.37 &  1/12 & 0h06  \\
    \bottomrule
    \\
    \end{tabular}
    }
    \caption{Test performance (top-1), the number of removed layers (Rem.) and training time for all the considered NLP setups.}
    \label{tab:NLP_training_time}
\end{table*}

\clearpage

\end{document}